\pdfoutput=1
\documentclass{article}

\PassOptionsToPackage{numbers, compress}{natbib}

\usepackage[preprint]{neurips_2023}
\usepackage{graphicx}

\usepackage[utf8]{inputenc} 
\usepackage[T1]{fontenc}    
\usepackage{hyperref}       
\usepackage{url}            
\usepackage{booktabs}       
\usepackage{amsfonts}       
\usepackage{nicefrac}       
\usepackage{microtype}      
\usepackage{xcolor}         
\usepackage{amsmath}        
\usepackage{diagbox}
\usepackage{caption} 
\usepackage{babel,blindtext}
\usepackage{wrapfig}
\usepackage{physics}
\usepackage{subcaption}

\title{An Empirical Study of Aegis}

\author{
    Daniel Saragih\thanks{Main correspondence} $^{, 1}$ \;
    Paridhi Goel$^{1}$ \;
    Tejas Balaji$^{1}$ \;
    Alyssa Li$^{1}$\vspace{0.7em}\\
    $^{1}$University of Toronto\vspace{.7em}\\
    \texttt{[daniel.saragih, paridhi.goel,}\\
    \hspace{0.5em}\texttt{tejas.balaji, alyssamengyuan.li]@mail.utoronto.ca}
}

\begin{document}
\maketitle

\begin{abstract}
    Bit flipping attacks are one class of attacks on neural networks with numerous defense mechanisms invented to mitigate its potency. Due to the importance of ensuring the robustness of these defense mechanisms, we perform an empirical study on the Aegis framework. We evaluate the baseline mechanisms of Aegis on low-entropy data (MNIST), and we evaluate a pre-trained model with the mechanisms fine-tuned on MNIST. We also compare the use of data augmentation to the robustness training of Aegis, and how Aegis performs under other adversarial attacks, such as the generation of adversarial examples. We find that both the dynamic-exit strategy and robustness training of Aegis has some drawbacks. In particular, we see drops in accuracy when testing on perturbed data, and on adversarial examples, as compared to baselines. Moreover, we found that the dynamic exit-strategy loses its uniformity when tested on simpler datasets \setcounter{footnote}{0}\footnote{Done as part of coursework for CSC413 at the University of Toronto.}.
    The code for this project is available on GitHub.
\footnote{\url{https://github.com/paridhi26/CSC413-project}}
\end{abstract}

\section{Introduction}

\begin{figure}[h]
    \centering
    \includegraphics[width=0.8\textwidth]{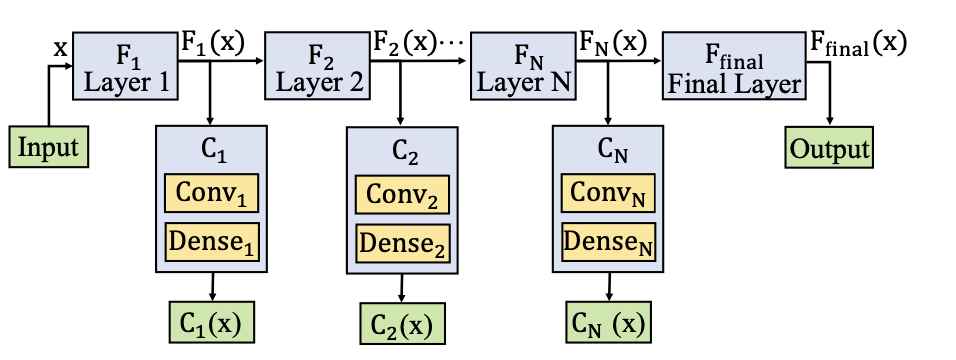}
    \caption{Base model (e.g. \textbf{Resnet32} or \textbf{VGG16}) equipped with internal classifiers $(C_1, \ldots, C_N)$ for early exiting; figure retrieved from the original paper \cite{wang2023aegis}.}
    \label{fig:sdn}
\end{figure}

The development of deep neural networks (DNN) has ushered a new era of security and safety applications, but along with it new vulnerabilities. One such concern is the threat of bit flipping attacks (BFAs), in which an adversary can tamper with critical model parameters, and negatively influence the accuracy of prediction. Recognizing the possibility of being attacked in this way, a defense mechanism called Aegis \cite{wang2023aegis}, applies a dynamic-exit strategy by attaching extra internal classifiers (ICs) to hidden layers (Figure \ref{fig:sdn}), while also performing robustness training (ROB) on the model by simulating BFAs during the fine-tuning of ICs (to be discussed in Section 2).

Motivated by the importance of mitigating such attacks on machine learning models, in this project, we conducted various tests on the Aegis framework. Specifically, our experiments were concerned with:
\begin{enumerate}
    \item Evaluating the test accuracy, accuracy on perturbed test data, and vulnerability to attack when using low-entropy data such as MNIST.
    \item Evaluating the test accuracy, accuracy on perturbed test data, and vulnerability to attack when trained on a general dataset like CIFAR10 and then fine-tuned\footnote{Fine-tuning, in our case, is slightly different from the conventional definition; see Section 2.} on MNIST.
    \item Evaluating the vulnerability of Aegis on other adversarial attacks such as the generation of adversarial examples \cite{goodfellow2015explaining}.
\end{enumerate}

\section{Background and Related Work}
The threat of BFAs on deep learning models rose to prominence with DeepHammer \cite{yao2020deep}, which aimed to reproduce the well-known Rowhammer attacks on DNNs. Defenses against the attack  generally fall into the categories of model augmentation or integrity verification \cite{wang2023aegis}; the defense that we will focus on falls in the former category. For instance, we have defenses such as BIN \cite{he2020defen} and RA-BNN \cite{rakin2021rabnn} which aims to binarize the model parameters, and in doing so increase the necessary bit flips to disrupt model performance. On the other hand, ModelShield \cite{guo2021model} uses hash verification to verify the integrity of model parameters. 

The Aegis paper proposes a defense mechanism which is "non intrusive, platform independent, and utility preserving" \cite{wang2023aegis}. The goal of the authors in developing a mechanism with these properties arose from the issues and difficulties encountered with previous defense mechanisms, such as having to retrain the entire model or having to change the parameters in the model. As such, Aegis only involves attaching ICs to the original model. Moreover, the training run only involves training the ICs while freezing the base model parameters; this procedure is what we refer to as \textit{fine-tuning}.

In light of its goal, Aegis adopts two defensive mechanisms. The first involves a dynamic-exit shallow-deep network (DESDN) where samples randomly exit the network early with a uniform probability distribution on the layers. Using the convention of Figure \ref{fig:sdn}, we sample $q$ candidates from a uniform distribution over $(C_1, \ldots, C_N)$. We then exit from the first of the candidates which exceed a certain confidence (defined as the highest probability over target classes) threshold $\tau$. Thus, when attackers attempt to target particular layers of the network, they will find it extremely difficult to pinpoint a set of layers that are likely to have the most negative effect on the prediction if attacked. This contrasts with a network without any early exists wherein the most vulnerable bits are those closest to the output classifier.

\begin{figure}[h]
    \centering
    \includegraphics[width=0.8\textwidth]{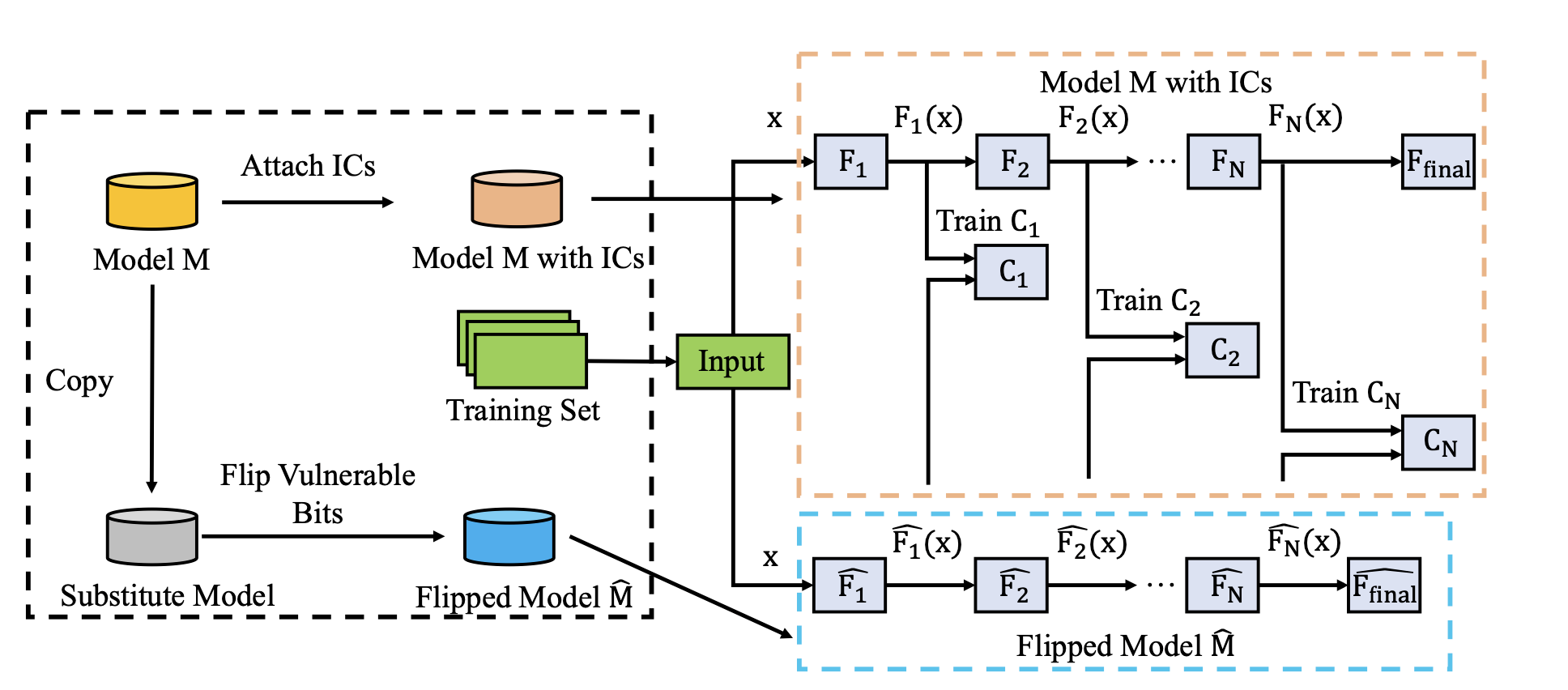}
    \caption{The ROB mechanism. We create a copy of the model, $\hat M$, referred to as the flipped model, and flip vulnerable bits of $\hat M$ to simulate an attack. Figure retrieved from the original paper \cite{wang2023aegis}.}
    \label{fig:rob}
\end{figure}

A second mechanism is robustness training (ROB) wherein the authors deliberately flipped vulnerable bits and trained the early exit classifiers using these features (Figure \ref{fig:rob}). To do so, we take a copy of the clean model $M$ to make a flipped model $\widehat M$. The vulnerable bits are identified through a gradient ranking algorithm. In particular, define $\mathcal{L} = L_{CE}(F_{final}(x), t)$ where $L_{CE}$ is the cross-entropy loss and $t$ is the ground-truth label. If we denote the parameter bits of $\widehat M$ as $B$, for each $b \in B$, calculate $\nabla_b \mathcal{L}$. Rank the gradients $\nabla_b \mathcal{L}$, and choose the top-k such bits to flip (where $k$ is the attack budget). During training we pass the training data and label through $M$ and $\widehat M$, and propagate the sum of their losses through the shared ICs.

\section{Methods}

\subsection{Base model training}
The focus of our methods is on studying the Aegis mechanisms. To start, as in the original paper, we used \textbf{ResNet32} \cite{he2015deep} and \textbf{VGG16} \cite{simonyan2015deep} as the base models. We trained four baseline models: \textbf{ResNet32} on MNIST and CIFAR10 \cite{kriz2012learn}, referred to as \textbf{R-MNIST} and \textbf{R-CIFAR} and \textbf{VGG16} on MNIST and CIFAR10: \textbf{V-MNIST}, \textbf{V-CIFAR}. 
In this initial training run, we froze the parameters of the internal classifiers $(C_1, \ldots, C_N)$ and instead made predictions using the main classifier head $F_{final}$. We ran the training loop for 30 epochs on MNIST and 60 epochs on CIFAR. We also used the Adam optimizer with a learning rate of $4 \times 10^{-4}$, and a batch size of 128. The output is an array of probabilities over the target classes and the loss used is cross-entropy.

\subsection{Fine-tuning}
Subsequently, we endowed these models with the Aegis ICs and further fine-tuned them. In particular, we fine-tuned three types of models: with ROB, without ROB, and with augmented data (without ROB). The augmentation strategy is our contribution motivated by the similarity of ROB to data augmentation strategies (i.e. both attempt to patch up under-represented areas of the learned distribution). The augmentation techniques we applied include rotation, translation, scaling, Gaussian blur, and random erasing. These model types will be distinguished with the suffix \textbf{-r}, \textbf{-nor}, and \textbf{-aug} respectively. In this fine-tuning session, we froze the base model parameters and only allowed gradients to change the internal classifier parameters. 

For the MNIST models, we fine-tuned with the same dataset. However, to accomplish our second objective, for both of our CIFAR models, \textbf{R-CIFAR} and \textbf{V-CIFAR}, we performed this fine-tuning step on the MNIST dataset. Thus, a model such as \textbf{R-CIFAR-r} is one initially trained on CIFAR and then fine-tuned on MNIST with ROB applied. The hyperparameters are the same as in base model training except we use 30 epochs when fine-tuning \textit{both} MNIST and CIFAR models. 


\subsection{Evaluation}
Prior work \cite{madry2019deep} has shown that training on adversarial samples in fact reduces the accuracy on real (non-adversarial) data as it dilutes the learned probability distribution. We highlighted this effect by performing inference on MNIST data which is relatively low entropy as compared to the datasets used in the original paper. In the same vein, we performed inference on perturbed data, where the perturbations are label-invariant (e.g. noising, blurring, and slight rotations). Overall, we made use of three types of evaluation procedures: 
\begin{enumerate}
    \item A baseline evaluation which tests on the fine-tuning dataset while using the early-exit strategy on fine-tuned models.
    \item A perturbed evaluation where label-invariant transformations are applied to the test data; again early-exit is used.
    \item An adversarial evaluation where we first attack the model with Proflip \cite{chen2021pro} and observe the attack success rate (ASR).
\end{enumerate}
In each case, we used the MNIST test set of $10000$ images.

An additional experiment that we performed aims to test Aegis' robustness to adversarial samples \cite{goodfellow2015explaining, papernot2016limitations}. As was shown in \cite{madry2019deep}, methods that directly focus on preventing specific attacks often suffer on other, unrelated attacks. We verified this finding on all of our fine-tuned models by reporting the accuracy on the samples produced by Fast Gradient Sign Methods (FGSM) \cite{goodfellow2015explaining} to judge how it fares against non-BFA adversaries. This is a rather simple attack that calculates the gradient with respect to the input data $\nabla_{\vb x}\mathcal{L}(\vb x, \vb t)$ and transforms the input in the direction of the gradient: \[
\vb{x}_{\text{perturbed}} = \vb x + \epsilon \text{sign} \nabla_{\vb x}\mathcal{L(\vb x, \vb t)} \quad \text{ for some $0 \leq \epsilon < 1$.}
\]

\section{Results}

\subsection{Baseline Evaluation}
Table \ref{table:evals} shows the test accuracies we got for the different baseline models when tested on the fine-tuning dataset while using the early-exit strategy on fine-tuned models. All \textbf{-r}, \textbf{-nor}, \textbf{-aug } were finetuned on MNIST and then were evaluated on MNIST. The basic non-finetuned models\textbf{ R-CIFAR} and \textbf{V-CIFAR} were trained on CIFAR and tested on CIFAR. Similarly, the non-finetuned models \textbf{R-MNIST} and \textbf{V-MNIST} were trained and tested on MNIST. Due to the ease of learning MNIST, a pattern is not apparent from the baseline accuracies alone, thus we subsequently consider the perturbed accuracies.
\begin{table}
    \centering
    \setlength\extrarowheight{3pt}
    \begin{tabular}{ccc}\toprule
        Model& Baseline Accuracy & Perturbed Accuracy  \\ \midrule
        R-CIFAR &  84.6 & 44.3  \\ 
        R-CIFAR-r & 97.8  & 60.6 \\ 
        R-CIFAR-nor & \textbf{98.9} & 70.3  \\ 
        R-CIFAR-aug & 98.7  & \textbf{92.0}  \\ \midrule
        V-CIFAR & 89.4  &  47.4 \\ 
        V-CIFAR-r &  90.2  &  40.6\\ 
        V-CIFAR-nor & \textbf{92.9} & 46.8 \\ 
        V-CIFAR-aug & 90.7  & \textbf{70.5} \\ \midrule
        R-MNIST &   98.3 & 72.7  \\ 
        R-MNIST-r & \textbf{99.3}  & 78.1 \\ 
        R-MNIST-nor & 98.7  & 65.4  \\ 
        R-MNIST-aug & 99.2  & \textbf{93.6}  \\ \midrule
        V-MNIST &  98.7 & 78.8 \\ 
        V-MNIST-r & 98.2 & 75.1 \\ 
        V-MNIST-nor & 98.9  & 79.9 \\ 
        V-MNIST-aug &  \textbf{99.3} & \textbf{92.1} \\ \bottomrule
        \end{tabular}
    \caption{Accuracies for all the models along with accuracies when tested on perturbed test data.}
    \label{table:evals}
\end{table}

\subsection{Perturbed Data Evaluation}
In this evaluation procedure, we perturbed the test data by adding some label-invariant transformations to it before testing. These transformations were random rotations, random translations, random scaling, Gaussian Blur, Gaussian noise and random erasing. 
We evaluated the models on perturbed test data to further pronounce the differences between different fine-tuning settings (\textbf{-r}, \textbf{-nor}, \textbf{-aug }). This is motivated by the fact that performance on perturbed test examples better demonstrates the ability of a model to learn general features of the data in a particular training set.

The perturbed accuracy column in Table \ref{table:evals} shows the test accuracies we got for the fine-tuned models (\textbf{-r}, \textbf{-nor}, \textbf{-aug }) when we tested them on the fine-tuning dataset with label-invariant transformations applied to the test data while using the early-exit strategy. It also shows the accuracies we got for the non-finetuned models (\textbf{R-MNIST}, \textbf{V-MNIST},  \textbf{R-CIFAR}, \textbf{V-CIFAR}) after similarly testing them on test data with label-invariant transformations applied. Excepting the perturbations, the training and evaluation datasets are the same as those used in Baseline Evaluation.

As expected, the two base (not fine-tuned) CIFAR models \textbf{ R-CIFAR} and \textbf{V-CIFAR}, showed the largest drop in accuracy when evaluated on perturbed CIFAR images compared to \textbf{R-MNIST} and \textbf{V-MNIST} which were evaluated on MNIST which is low entropy. CIFAR images are more complex than MNIST images and perturbation in test data generally affected all CIFAR models more compared to their MNIST counterparts. The only outlier here was \textbf{R-MNIST-nor} which showed lower than expected Perturbed accuracy.

We also expected that the \textbf{-r} models which went through robustness training during the fine tuning process to perform worse than the \textbf{-nor} models which did not go through robustness training since robustness training can dilute the learned probability distribution. This hypothesis was also supported by our evaluation results, with \textbf{R-MNIST-nor} being the only exception as above.

For the \textbf{-aug} fine-tuning setting, we added some of the label-invariant transformations we used to perturb test data to the fine-tuning data. For \textbf{-aug}, the transformation that were applied to the training data while fine tuning were random rotations, random translations, random scaling, Gaussian Blur and random erasing. That is all the transformations from the perturbed test data except Gaussian noise. Since, these models where already fine tuned on perturbed data, their accuracy drop was least significant on the perturbed test data. 

\subsection{Proflip}

In this section, we present results for the Proflip \cite{chen2021pro} evaluations. The Proflip procedure inserts a backdoor into the target model by flipping parameter bits such that the prediction of all inputs with the set trigger will result in a pre-specified, incorrect target class. The ASR of Proflip is thus the proportion of trigger examples that manage to fool the model into predicting incorrectly. The paper presents two variants of Proflip: adaptive and non-adaptive \cite{chen2021pro}; we will limit ourselves to the latter.

\begin{wraptable}{r}{8cm}
\caption{Final ASR values for \textbf{-r}, \textbf{-nor}, and \textbf{-aug} models on Proflip}
\label{table:asr_proflip}
\begin{tabular}{cc}
\toprule
Model & ASR\\\midrule
R-MNIST-r   & 22.4 \\
R-MNIST-nor  & 34.9\\
R-MNIST-aug  & \textbf{11.5}\\\midrule
V-MNIST-r   & 14.5 \\
V-MNIST-nor  & \textbf{11.6}\\
V-MNIST-aug  & 12.0\\\bottomrule
\end{tabular}\quad
\begin{tabular}{cc}
\toprule
Model & ASR\\\midrule
R-CIFAR-r   & 13.4 \\
R-CIFAR-nor  & 14.9 \\
R-CIFAR-aug  & \textbf{12.7}\\ \midrule
V-CIFAR-r   & 14.4 \\
V-CIFAR-nor  & 15.2 \\
V-CIFAR-aug  & \textbf{10.6}\\\bottomrule
\end{tabular}\quad
\end{wraptable}

\begin{figure}
    \centering
    \begin{subfigure}[b]{0.48\textwidth}
        \caption{}
        \includegraphics[width=\textwidth]{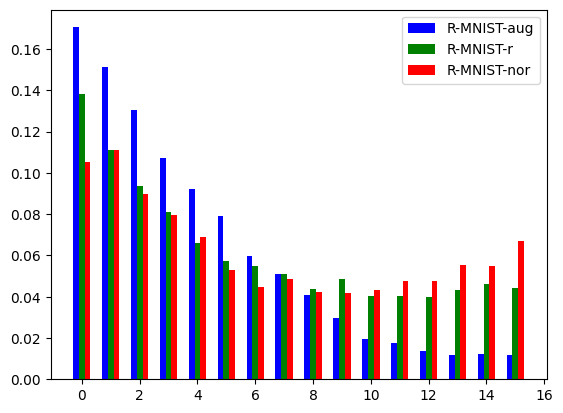}
        \label{fig:pro_mnist_res}
    \end{subfigure}
    \begin{subfigure}[b]{0.48\textwidth}
        \caption{}
        \includegraphics[width=\textwidth]{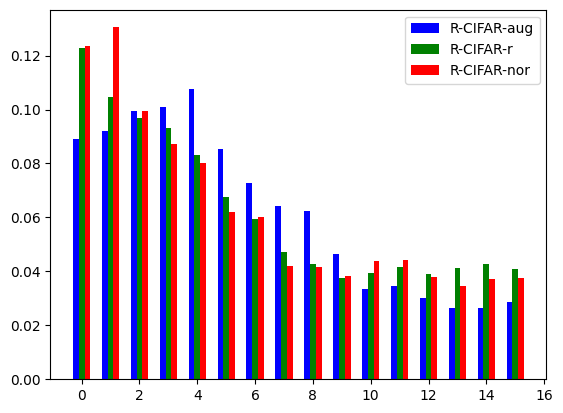}
        \label{fig:pro_cifar_res}
    \end{subfigure}
    \caption{Number of exits per layer for \textbf{R-MNIST} and \textbf{R-CIFAR} models}
    \label{fig:proflip_res}
\end{figure}

\begin{figure}
    \centering
    \begin{subfigure}[b]{0.48\textwidth}
        \caption{}
        \includegraphics[width=\textwidth]{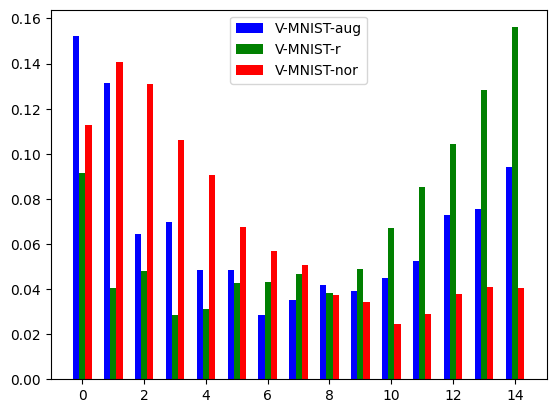}
        \label{fig:pro_mnist_vgg}
    \end{subfigure}
    \begin{subfigure}[b]{0.48\textwidth}
        \caption{}
        \includegraphics[width=\textwidth]{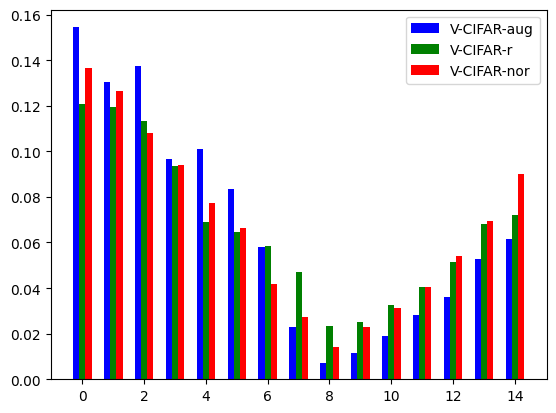}
        \label{fig:pro_cifar_vgg}
    \end{subfigure}
    \caption{Number of exits per layer for \textbf{V-MNIST} and \textbf{V-CIFAR} models}
    \label{fig:proflip_vgg}
\end{figure}

Table \ref{table:asr_proflip} shows the final ASR for all the models which we attacked with ProFlip (using an attack budget of $N_b = 50$). As the results indicate, the augmented models tend to have the lowest ASR values (with the exception of the \textbf{V-MNIST} models). The next best class of models tends to be the \textbf{-r} models, while the \textbf{-nor} models are observed to have the highest ASR. Moreover, we tracked the number of exits made in the IC layers for each of the models when attacked with ProFlip. In general, we observed that the Resnet models seemed to have more exits in earlier layers as compared to the VGG models. The results for number of exits per layer for \textbf{R-MNIST} and \textbf{V-MNIST} are shown in Figures \ref{fig:proflip_res}  and \ref{fig:proflip_vgg}.

We expected the -\textbf{nor} models to perform the worst out of all the three classes (\textbf{-r}, \textbf{-aug}, and \textbf{-nor}) since they have been trained with the least amount of resistance to bit flip attacks. However, what we do find surprising is that the \textbf{-aug} models perform better than the \textbf{-r} models in most cases. This confirms our hypothesis that augmentation serves a similar role to ROB. The improvement seen in the case of augmentation likely stems from the fact that label-invariant transformations are more informative for learning than the outputs of a flipped model. This increased generality would allow the model to fare better when it encounters an unfamiliar attack pattern. It is also interesting to note the ASR difference between \textbf{R-MNIST} and \textbf{R-CIFAR} models. This indicates that the more general features learned through fine-tuning makes the model more robust. However, the same cannot be said of the VGG models, and so more testing is necessary to verify this trend.

As is depicted in Figures \ref{fig:proflip_res} and \ref{fig:proflip_vgg}, the \textbf{R-MNIST} and \textbf{R-CIFAR} models tend to have a lower proportion of exits occurring in later layers, as compared to the \textbf{V-MNIST} and \textbf{V-CIFAR} models. We believe that one of the causes for this is that ResNet, owing to its name, uses residual connections as compared to VGG, which suffers from a vanishing gradients issue. MNIST being a low-entropy and "easy" dataset means that strong classifiers such as Resnet will be able to easily obtain very high accuracies within just a few layers. On the other hand, the vanishing gradients issue with VGG is likely to affect the amount of layers it takes for VGG to classify MNIST images with high confidence (as compared to Resnet). 

It is also interesting to note the skew towards earlier layers. As mentioned in the methods, DESDN randomly picks a set of $q$ candidate classifiers and then early exits when it encounters the first classifier that achieves an accuracy above a specific threshold. Thus, it is likely that there is a bias towards the earlier layers, given that MNIST is an easy dataset to learn. If the modelling problem was in fact more challenging, it's likely that later layers are required to exceed the confidence threshold. In turn, this would provide a counter balance to the early-exit bias. Thus, we may conclude that the uniformity in early-exiting which Aegis shows in the original paper \cite{wang2023aegis} is in fact contingent on a fragile balancing act between the early-exit bias, and the requirement of later layers to achieve high confidence on predictions.


\subsection{Adversarial Examples}

\begin{table}
    \centering
    \begin{tabular}{ l c c c c c }\toprule
        \diagbox[width=10em]{Model}{$\epsilon$}&
          0.00 & 0.05 & 0.10 & 0.15 & 0.20 \\ \midrule
        R-CIFAR & 73.8  & 32.0 & 16.7 & 11.1 & 9.14 \\ 
        R-CIFAR-r & 99.0  & 61.5 & 24.6 & 14.9 & 11.6\\ 
        R-CIFAR-nor & \textbf{99.4} & 89.7 & 49.3 & 19.2 & 13.5  \\ 
        R-CIFAR-aug & 99.3  & \textbf{92.7} & \textbf{82.0} & \textbf{62.3} & \textbf{37.6} \\ \midrule 
        V-CIFAR & 81.6  & 35.3 & 14.6 & 9.28 & 8.55 \\ 
        V-CIFAR-r & 97.5  & 48.0 & 12.3 & 10.1 & \textbf{9.98} \\ 
        V-CIFAR-nor & \textbf{99.2}  & 75.0 & 19.9 & 10.3 & 9.80\\ 
        V-CIFAR-aug & 99.0  & \textbf{91.3} &\textbf{ 51.1} & \textbf{13.8} & 9.82\\ \midrule 
        R-MNIST & 98.2  & 96.7 & \textbf{94.7} & 73.8 & 36.9 \\ 
        R-MNIST-r & 99.5  & 97.2 & 86.0 & 37.0 & 10.1\\ 
        R-MNIST-nor & 99.0  & 94.3 & 59.0 & 17.6 & 10.3  \\ 
        R-MNIST-aug & \textbf{99.5}  & \textbf{97.3} & 93.4 & \textbf{86.9} & \textbf{73.1} \\ \midrule 
        V-MNIST & 98.7  & \textbf{96.8} & \textbf{93.3} & 72.0 & 41.7\\ 
        V-MNIST-r & 98.9 & 93.7 & 76.4 & 41.4 & 19.1\\ 
        V-MNIST-nor & \textbf{99.0}  & 93.8 & 70.0 & 26.1 & 13.7\\ 
        V-MNIST-aug & 95.1  & 88.5 & 87.4 & \textbf{80.5} & \textbf{46.9}\\ \bottomrule
        \end{tabular}
    \caption{Accuracies for all the models when tested on adversarial examples. It should be noted again that R/V-CIFAR is evaluated on the CIFAR10 dataset as opposed to MNIST.}
    \label{table:adv_ex}
\end{table}

Table \ref{table:adv_ex} shows the accuracies of the models we trained against various values of $\epsilon$, which denotes the strength of perturbation on input data in FGSM attacks. We observe that the accuracies for the Resnet augmented models tend to be the highest among all other Resnet models. For the CIFAR models, we see that for $\epsilon = 0.2$, the \textbf{R-CIFAR-aug} model achieves an accuracy of 37.6\% while \textbf{R-CIFAR-nor} and \textbf{R-CIFAR-r} achieve less than 15\% accuracy. The same goes for the MNIST {Resnet} models, where \textbf{R-MNIST-aug} achieves a very high accuracy of 73.1\% on $\epsilon = 0.2$ while the \textbf{-r} and \textbf{-nor} models achieve accuracies below 20\%. As for the VGG models, the difference in accuracy doesn't seem to be very high across the models for higher values of $\epsilon$ in the case of CIFAR. However, in the case of MNIST, \textbf{V-MNIST-aug} seems to greatly outperform its \textbf{-r} and \textbf{-nor} counterparts, achieving an accuracy of 46.9\% for $\epsilon = 0.2$. We observe moreover that the baseline models tend to outperform the \textbf{-r} and \textbf{-nor} models, and are (in most cases) second only to the augmented models \footnote{We should also note the caveat that \textbf{R/V-CIFAR} models are tested on CIFAR10.}.  

\begin{figure}
    \centering
    \begin{subfigure}[b]{0.48\textwidth}
        \caption{}
        \includegraphics[width=\textwidth]{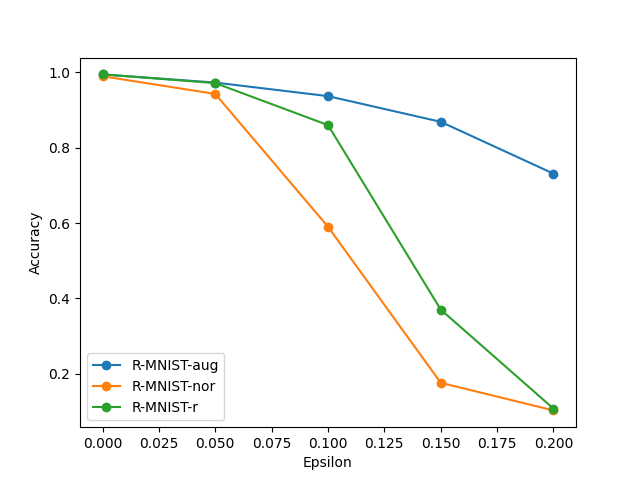}
        \label{fig:adv_mnist_res}
    \end{subfigure}
    \begin{subfigure}[b]{0.48\textwidth}
        \caption{}
        \includegraphics[width=\textwidth]{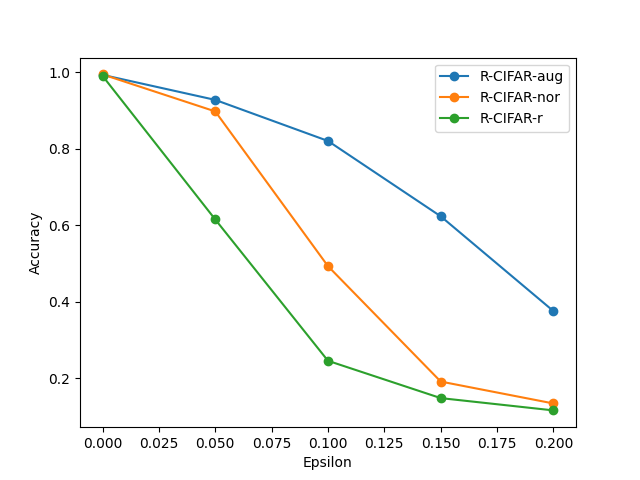}
        \label{fig:adv_mnist_vgg}
    \end{subfigure}
    \caption{Epsilon vs Accuracy for fine-tuned \textbf{R-MNIST} and \textbf{R-CIFAR} models.}
    \label{fig:adv_resnet}
\end{figure}

\begin{figure}
    \centering
    \begin{subfigure}[b]{0.48\textwidth}
        \caption{}
        \includegraphics[width=\textwidth]{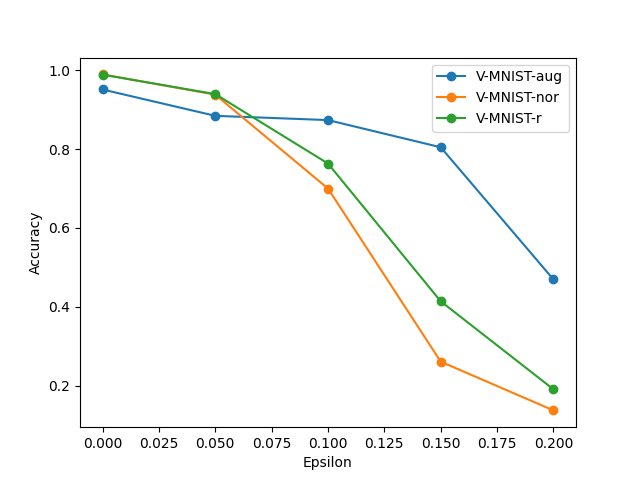}
        \label{fig:adv_cifar_res}
    \end{subfigure}
    \begin{subfigure}[b]{0.48\textwidth}
        \caption{}
        \includegraphics[width=\textwidth]{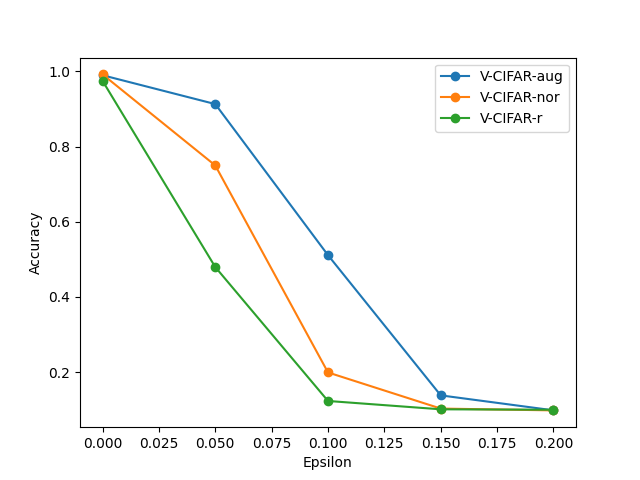}
        \label{fig:adv_cifar_vgg}
    \end{subfigure}
    \caption{Epsilon vs Accuracy for fine-tuned \textbf{V-MNIST} and \textbf{V-CIFAR} models.}
    \label{fig:adv_vgg}
\end{figure}

Let a "group" of models denote one the four groups of 4 models as shown in Table \ref{table:adv_ex} (for example, the first group is the group of all \textbf{R-CIFAR} models). We observe that in 3 out of the 4 groups, the augmented model vastly outperformed its baseline, \textbf{-r}, and \textbf{-nor} counterparts. It was only in the case of \textbf{V-CIFAR} that all models were observed to have a similar accuracy for $\epsilon = 0.2$. Barring the outlier for \textbf{V-CIFAR}, we claim that the results we obtained are to be expected, for the following reasons:
\begin{enumerate}
    \item The \textbf{-nor} models performing worse than the \textbf{-aug} models is expected since the \textbf{-aug} models have been trained with perturbations of the data in mind. The \textbf{-aug} models have been trained to correctly classify noisy and perturbed data, making them robust to a wide range of adversarial inputs. Moreover, the baseline models were expected to outperform the \textbf{-nor} and \textbf{-r} models, since the latter two models have been trained to specifically combat bit-flip attacks. In particular, we note that both the \textbf{-nor} and \textbf{-r} have been fine-tuned on the baseline models for the purpose of preventing bit-flip attacks. This in turn changes the parameters and weights in the classifiers. Thus, a significant portion of robustness that the baseline models possessed to adversarial data was lost when the models were fine-tuned. In the case of the \textbf{-aug} models, although they were trained on top of the \textbf{-nor} models and not the baseline, they regained this robustness as a result of the augmented training. We also observe that in two of the four groups, the \textbf{-nor} models outperform the \textbf{-r} models. In particular, this is observed to be the case in the \textbf{R-CIFAR} and \textbf{V-CIFAR} model groups. Thus, this unusual behavior seems to be a result of differences between the MNIST and CIFAR datasets and not a result of the baseline models (i.e., VGG and Resnet). We hypothesize that this is the result of MNIST being a low-entropy dataset as compared to CIFAR, and thus easier to classify.
    \item The most intriguing result is that the \textbf{-aug} models perform significantly better than the \textbf{-r} models. Figures \ref{fig:adv_resnet} and \ref{fig:adv_vgg} showcase this stark difference in accuracies between the two models. While the \textbf{-r} models have been through robustness training, the training has specifically been for the prevention of bit-flip attacks. The use of adversarial examples is a different mechanism of attacking a neural network, it is thus expected that a model that has not built any resistance for this specific type of attack will prove to be vulnerable. The augmented models, however, are more general in the class of perturbations (and hence, attacks) they can resist. 
\end{enumerate}

\section{Conclusion, Limitations, and Future Work}

The results presented indicate that the defensive framework proposed in Aegis possesses significant defensive capabilities as shown by low ASR in Table \ref{table:asr_proflip}, but also some notable drawbacks. From our experiments, we observed that ROB often stunts the model's ability to learn with sufficient generality. For instance, in Table \ref{table:evals}, we see that \textbf{-r} models generally performed worse than its counterparts. Moreover, the protection ROB provides against BFAs comes at the cost of other adversarial attacks such as FGSM, as shown by its diminished accuracy in Table \ref{table:adv_ex}. In addition, DESDN also has its downsides. For instance, just by evaluating on low-entropy datasets such as MNIST, we are able to obtain a non-uniform exit distribution, contrary to the uniformity showcased in the original paper. This suggests that if an attacker is aware of the type of datasets a model is tested on, they will have an easier time attacking the model.

Our work also has its limitations. For instance, we claim in Section 4.3 that there is a balancing act between the number of layers needed to achieve the confidence threshold and the early-exit bias of DESDN. Although our experiments showcase this effect, more testing on different datasets is required to verify its validity. Therefore, future work should focus on performing further verification of the effects hinted at by our work. An interesting avenue involves conducting a wider array of adversarial attacks on Aegis, or using new methods to generate adversarial samples. And certainly, future work should also be dedicated to finding improvements to Aegis that mitigates some of the issues brought up in our work.

\medskip
{
\small
\bibliographystyle{unsrt}
\bibliography{report_final}

\begin{thebibliography}{10}

\bibitem{wang2023aegis}
Jialai Wang, Ziyuan Zhang, Meiqi Wang, Han Qiu, Tianwei Zhang, Qi~Li, Zongpeng
  Li, Tao Wei, and Chao Zhang.
\newblock Aegis: Mitigating targeted bit-flip attacks against deep neural
  networks, 2023.

\bibitem{goodfellow2015explaining}
Ian~J. Goodfellow, Jonathon Shlens, and Christian Szegedy.
\newblock Explaining and harnessing adversarial examples, 2015.

\bibitem{yao2020deep}
Fan Yao, Adnan~Siraj Rakin, and Deliang Fan.
\newblock {DeepHammer}: Depleting the intelligence of deep neural networks
  through targeted chain of bit flips.
\newblock In {\em 29th USENIX Security Symposium (USENIX Security 20)}, pages
  1463--1480. USENIX Association, August 2020.

\bibitem{he2020defen}
Zhezhi He, Adnan~Siraj Rakin, Jingtao Li, Chaitali Chakrabarti, and Deliang
  Fan.
\newblock Defending and harnessing the bit-flip based adversarial weight
  attack.
\newblock In {\em 2020 IEEE/CVF Conference on Computer Vision and Pattern
  Recognition (CVPR)}, pages 14083--14091, 2020.

\bibitem{rakin2021rabnn}
Adnan~Siraj Rakin, Li~Yang, Jingtao Li, Fan Yao, Chaitali Chakrabarti, Yu~Cao,
  Jae sun Seo, and Deliang Fan.
\newblock Ra-bnn: Constructing robust \& accurate binary neural network to
  simultaneously defend adversarial bit-flip attack and improve accuracy, 2021.

\bibitem{guo2021model}
Yanan Guo, Liang Liu, Yueqiang Cheng, Youtao Zhang, and Jun Yang.
\newblock Modelshield: A generic and portable framework extension for defending
  bit-flip based adversarial weight attacks.
\newblock In {\em 2021 IEEE 39th International Conference on Computer Design
  (ICCD)}, pages 559--562, 2021.

\bibitem{he2015deep}
Kaiming He, Xiangyu Zhang, Shaoqing Ren, and Jian Sun.
\newblock Deep residual learning for image recognition, 2015.

\bibitem{simonyan2015deep}
Karen Simonyan and Andrew Zisserman.
\newblock Very deep convolutional networks for large-scale image recognition,
  2015.

\bibitem{kriz2012learn}
Alex Krizhevsky.
\newblock Learning multiple layers of features from tiny images.
\newblock {\em University of Toronto}, 2012.

\bibitem{madry2019deep}
Aleksander Madry, Aleksandar Makelov, Ludwig Schmidt, Dimitris Tsipras, and
  Adrian Vladu.
\newblock Towards deep learning models resistant to adversarial attacks, 2019.

\bibitem{chen2021pro}
Huili Chen, Cheng Fu, Jishen Zhao, and Farinaz Koushanfar.
\newblock Proflip: Targeted trojan attack with progressive bit flips.
\newblock In {\em 2021 IEEE/CVF International Conference on Computer Vision
  (ICCV)}, pages 7698--7707, 2021.

\bibitem{papernot2016limitations}
Nicolas Papernot, Patrick McDaniel, Somesh Jha, Matt Fredrikson, Z.~Berkay
  Celik, and Ananthram Swami.
\newblock The limitations of deep learning in adversarial settings.
\newblock In {\em 2016 IEEE European Symposium on Security and Privacy
  (EuroS\&P)}, pages 372--387, 2016.

\end{thebibliography}

}


\end{document}